\documentclass[letterpaper, 10 pt, conference]{ieeeconf}  

\IEEEoverridecommandlockouts                              

\usepackage{amsmath} 
\usepackage{amssymb}  

\usepackage{hyperref} 
\usepackage{graphicx}
\usepackage[subrefformat=parens]{subcaption}
\usepackage{booktabs}
\usepackage{balance}
\usepackage{algorithm}
\usepackage{algpseudocode}
\usepackage{adjustbox}

\title{\LARGE \bf
Gaze-Guided Task Decomposition \\ for Imitation Learning in Robotic Manipulation
}

\author{Ryo Takizawa$^{*1}$, Yoshiyuki Ohmura$^{1}$, Yasuo Kuniyoshi$^{1}$
\thanks{$^{1}$Laboratory for Intelligent Systems and Informatics, Graduate School of Information Science and Technology,
The University of Tokyo, 7-3-1 Hongo, Bunkyo-ku, Tokyo, Japan 
(Email: \tt\small\{takizawa, ohmura, kuniyosh\}@isi.imi.i.u-tokyo.ac.jp)}%
\thanks{$^{*}$Corresponding Author}%
}

\begin{document}

\maketitle
\thispagestyle{empty}
\pagestyle{empty}

\begin{abstract}
In imitation learning for robotic manipulation, decomposing object manipulation tasks into sub-tasks enables the reuse of learned skills and the combination of learned behaviors to perform novel tasks, rather than simply replicating demonstrated motions. 
Human gaze is closely linked to hand movements during object manipulation.
We hypothesize that an imitating agent's gaze control—fixating on specific landmarks and transitioning between them—simultaneously segments demonstrated manipulations into sub-tasks. 
This study proposes a simple yet robust task decomposition method based on gaze transitions. 
Using teleoperation, a common modality in robotic manipulation for collecting demonstrations, in which a human operator's gaze is measured and used for task decomposition as a substitute for an imitating agent's gaze.
Our approach ensures consistent task decomposition across all demonstrations for each task, which is desirable in contexts such as machine learning. 
We evaluated the method across demonstrations of various tasks, assessing the characteristics and consistency of the resulting sub-tasks. Furthermore, extensive testing across different hyperparameter settings confirmed its robustness, making it adaptable to diverse robotic systems.
Our code is available at \href{https://github.com/crumbyRobotics/GazeTaskDecomp}{https://github.com/crumbyRobotics/GazeTaskDecomp}.
\end{abstract}

\section{INTRODUCTION}
\begin{figure}
    \begin{tabular}{c}
         \begin{minipage}[b]{0.5\linewidth}                     
            \centering
            \includegraphics[width=\linewidth]{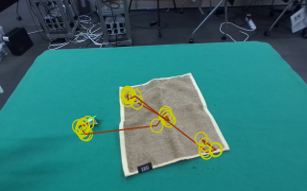}
            \subcaption{}
            \label{gaze-structure-a}
         \end{minipage}
         \\
         \begin{minipage}[c]{0.95\linewidth}                     
            \centering
            \includegraphics[width=\linewidth]{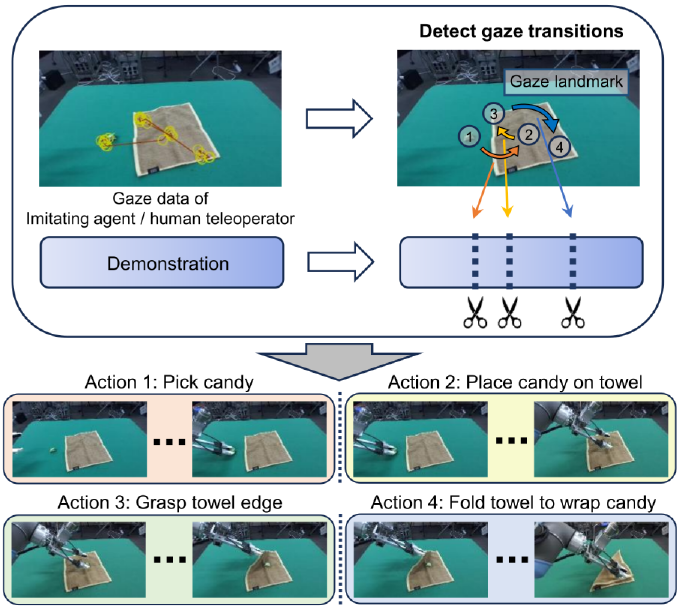}
            \subcaption{}
            \label{gaze-structure-c}
         \end{minipage}
    \end{tabular}
    \caption{Gaze itself possesses an inherent structure in object manipulation, which enables task decomposition by detecting transitions between gaze landmarks during demonstration. (a) Gaze transition of a human teleoperating a robot.
    (b) Overview of our proposed method.}
    \label{gaze-structure}
\end{figure}
   
Imitation is a promising approach for enabling robots to function as autonomous agents capable of acquiring object manipulation skills that align with human intentions. 
The ability of an imitating agent to automatically decompose a demonstrated manipulation into meaningful sub-tasks is a key challenge. This capability is essential for executing long-horizon tasks and for reusing or combining learned skills.

An effective method is required to segment demonstrations of various tasks into sub-tasks without the need for additional training or diverse data collection.
Existing deep learning-based imitation learning methods typically demand hundreds of demonstrations per task, making it crucial to ensure consistent sub-task decomposition across all demonstrations. 
However, few methods have been proposed for such task decomposition, and none achieve consistent task segmentation.
As a result, demonstrations are often manually segmented into sub-tasks \cite{Kim2024}, otherwise the goal state, which conditions a policy model, is usually substituted with a state a certain number of steps ahead \cite{Wang2023, Black2024}.

This study focuses on gaze information to achieve consistent task decomposition without the need for additional training or diverse data collection.
As Yarbus stated, “Eye movement reflects the human thought processes; so the observer’s thought may be followed to some extent from records of eye movement” \cite{Yarbus1967}, highlighting gaze as a naturally evolved tool for understanding ongoing events.
During human object manipulations, individuals repeatedly fixate on task-relevant landmarks before shifting to the next (Fig. \ref{gaze-structure-a}). These gaze transition behaviors are closely coupled with the motor planning processes of hands \cite{Hayhoe2005,Land1999, Hayhoe2003}.
These insights suggest that leveraging an imitating agent's gaze information, including its position and associated visual features, serves as an effective approach for task decomposition in demonstrated object manipulations.

In this paper, we demonstrate that detecting transitions between gaze landmarks during imitation enables robust and consistent task decomposition of demonstrations (Fig. \ref{gaze-structure-c}).
In the imitation learning framework, demonstrations are commonly collected via human teleoperation. 
Several approaches have incorporated simultaneous gaze tracking alongside motion recording, using the teleoperator’s gaze as a proxy for an imitating agent’s gaze \cite{Kim2020, Kim2021}.
These methods have enhanced the robustness and dexterity of the imitated actions with minimal modifications to the conventional teleoperation systems.
Building on this concept, we record the teleoperator’s gaze information (Fig. \ref{gaze-structure-a}) and use it for task decomposition.
Our proposed method segments demonstrations based on gaze transition timings, detected through a threshold-based approach that identifies abrupt changes in gaze position and visual features.
To ensure consistent task decomposition across all demonstrations of the same task, we introduce a refinement process that automatically adjusts these thresholds for each demonstration.

The application of this method is demonstrated using datasets of three tasks collected for imitation learning and evaluated the characteristics and consistency of the resulting sub-tasks. 
Furthermore, an investigation of the proposed refinement process in improving the consistency of task decomposition and the robustness to variations in hyperparameters is carried out.

\section{RELATED WORK}
\subsection{Robot Manipulation Imitation Using Task Decomposition}
Several proposed approaches have used temporal task decomposition for imitation learning in robot manipulation. 
A common approach for enhancing robustness in long-horizon tasks is the use of goal-conditioned policy models \cite{Kim2024, Wang2023}. 
The decomposition of tasks into appropriate sub-tasks and presenting them as sequential goals facilitates more effective execution.
Recently, language-conditioned models aimed at multi-task capabilities have gained popularity \cite{Brohan2022, Bharadhwaj2023, Kim2024b, Moo2024}. 
In datasets for language-conditioned models, sub-task-level instructions, such as "open the drawer" and "place the toy in the drawer," are generally preferred over task-level instructions, such as "put the toy in the drawer" \cite{Myers2024}. 
This suggests that goal- or language-conditioned methods inherently assume that object manipulations in the given demonstrations are already segmented into individual sub-tasks.
However, in most existing work, this segmentation of these demonstrations is performed manually. Our proposed method addresses this limitation by automating part of the data collection process, reducing annotation costs and streamlining the data collection process.

Recent imitation learning approaches often collect demonstrations via remote control. To enhance operability, teleoperators commonly use head-mounted displays (HMDs) that provide a live video feed using onboard cameras \cite{Zhang2018, Seo2023, Cheng2024, Iyer2024}. Our proposed task decomposition method leverages the teleoperator’s gaze data to guide the decomposition process. With advancements in gaze-tracking technology, HMDs with integrated gaze tracking are now commercially available. This enables our approach to be easily applied across various robotic systems by simply replacing the HMD used during remote operation.

\subsection{Conventional Task Decomposition Methods}
In \textit{Action Segmentation}, numerous methods have been proposed for the temporal segmentation of actions and object manipulation motions \cite{Gupta2009, Hyder2024, Dreher2020, Worgotter2012, Lea2016, Fathi2011, Huang2016}, including approaches that incorporate human gaze information \cite{Hipiny2017, Fathi2012}. However, most of these methods are designed for human action recognition, and very few are suitable for task decomposition in robotic object manipulation. Moreover, many rely heavily on visual, data-driven techniques, requiring new datasets tailored to each robotic platform for effective deployment.

Several studies have explored task decomposition in robotic object manipulation. For instance, Liang et al. optimized transitions between the manipulating (tool) and manipulated (target) objects to segment tasks, though their method requires the 6D poses of all relevant objects \cite{Liang2022}. Lee et al. reduced the dimensionality of whole-body trajectories using PCA and modeled them with GMM, assigning each Gaussian component to a sub-task \cite{Lee2015}. Zhu et al. employed neural networks for latent-space clustering to achieve task decomposition \cite{Zhu2022}. Zhang et al. embedded entire images using pre-trained models and detected sub-task transitions based on the distances between the resulting embeddings \cite{Zhang2024}.
However, these approaches have several limitations. Liang et al.'s method demands the 6D poses of all relevant objects, while Lee et al.'s approach depends on hand motion data, making it sensitive to variations in such movements. Zhu et al.'s technique requires retraining the neural network for new tasks and determines the number of sub-tasks based solely on time step lengths. Although Zhang et al.'s method enables detection-based task decomposition without additional data or training, it relies on the distance metrics in the embedding space of entire images, which may not reliably capture the task structure. 

In contrast, our proposed method addresses these shortcomings by eliminating the need for explicit environmental modeling, not relying on hand motion data, avoiding the training of neural networks, and determining the number of sub-tasks based on the number of gaze transitions.
Furthermore, compared to distance metrics in the embedding space of the entire images, gaze transitions have been extensively studied for their coordination with hand movement planning in object manipulation, making our gaze-guided approach more reliable.

\section{METHOD}
\label{sec:section3}

This section defines task decomposition of object manipulation based on gaze transitions, followed by a detailed implementation procedure. 
We begin by explaining how gaze transitions are detected from time-series gaze data, and subsequently describe the refinement process used to achieve consistent decomposition across the demonstration dataset.

\subsection{Gaze-based Task Decomposition}
A scenario is considered in which a human remotely controls a robot using only visual information to collect demonstrations of object manipulation for imitation learning.
Multiple demonstrations (typically ranging from several tens to hundreds) are collected for each task, with all demonstrations of the same task sharing an identical set of sub-tasks.
Let $\{g_t\}_{t=0}^{T}$ denote the time-series of the teleoperator’s gaze positions and $\{f_t\}_{t=0}^{T}$ the time-series of the visual features observed around those gaze positions. 
Given that the demonstrated task is decomposed into a set of sub-tasks $\{\tau_k\}_{k=0}^{K}$, the objective is to determine the $K-1$ time steps at which the sub-tasks switch (i.e. $\{t \mid k(t+1) = k(t) + 1\}$), using the gaze information $\{g_t\}$ and $\{f_t\}$.

In human object manipulation, each task is known to have a specific set of gaze landmarks $\{l_i\}_{i=0}^{N}$, where humans fix their gaze on each landmark for a certain duration before sequentially transitioning to the next landmark  \cite{Land1999, Hayhoe2003, Hayhoe2005}. The landmark index $i$ is arranged according to the order in which landmarks are fixated from the start of the task (duplicates may occur). 

In this study, we hypothesize that the time steps at which the index of the fixated gaze landmark changes, $\{t \mid i(t+1) = i(t)+1\}$, correspond to the time steps at which the sub-tasks switch, $\{t \mid k(t+1) = k(t) + 1\}$. In other words, task decomposition can be achieved by detecting gaze landmark transitions.

Let $p^{i(t)}_t$ be the location of the landmark $l_{i(t)}$ in the teleoperator's field of view at time step $t$. 
Then, the gaze position $g_t$ and the visual feature $f_t$ at that gaze position can be written as follows:
\begin{equation}
    \label{3_g_t}
    g_t = p^{i(t)}_t + \epsilon_t,
\end{equation}
\begin{equation}
    \label{3_f_t}
    f_t = f\bigl(g_t\bigr) + \zeta_t = f\bigl(p^{i(t)}_t + \epsilon_t\bigr) + \zeta_t,
\end{equation}
where $\epsilon_t$ and $\zeta_t$ represent noise in the gaze position and perturbations in the visual features caused, for example, by the end-effector movements, respectively, and $f(\cdot)$ is the visual feature extractor at the given gaze position.

By leveraging the relationships in Eqs. \ref{3_g_t} and \ref{3_f_t}, we can detect the transitions of gaze landmarks $\{l_i\}$ from the gaze information $\{g_t\}$ and $\{f_t\}$, enabling the task decomposition of the demonstrated object manipulation.

\subsection{Gaze Transition Detection}
\label{gaze-transition-detection}
The implementation details for segmenting demonstrations based on gaze transitions are described. As discussed in the previous section, detecting the time steps at which gaze landmarks change using time-series gaze data is essential for task decomposition.

\subsubsection{Median Filtering}
Gaze data typically contain noise. Even when the teleoperator is fixating on a stationary gaze landmark, recorded gaze positions may fluctuate. Moreover, brief shifts in gaze to different landmarks can occur during object manipulation. To address this noise effectively, we adopt a median-filtering approach. Specifically, for the teleoperator’s gaze position $g_t$, we apply the process shown in Eq. \ref{3_median}, with a window size $w$ as a hyperparameter. This resulting filtered gaze data $\hat{g}_t$ is then used in all subsequent processing:
\begin{align}
    \label{3_median}
    \hat{g}_t = \text{median}\bigl(g_{t-\tfrac{w}{2}:t+\tfrac{w}{2}}\bigr).
\end{align}

\subsubsection{Visual Features Around the Gaze}
Gaze transitions cause not only positional changes in the gaze but also variations in surrounding visual features. 
This study also considers detecting gaze transitions using visual features.
Accurately identifying changes in visual features requires a feature extractor built on high-quality representation learning.
For each time step, we crop an $b \times b$ region centered at the filtered gaze position $\hat{g}_t$ from both the left and right images. 
These cropped images are then processed through a pre-trained CLIP model \cite{Radford2021} to extract their feature vectors $f_t \in \mathbb{R}^{512 \times 2}$. 
This time-series of feature vectors, along with the gaze position, is used for gaze transition detection in this study.

\begin{figure}
    \includegraphics[width=0.95\linewidth]{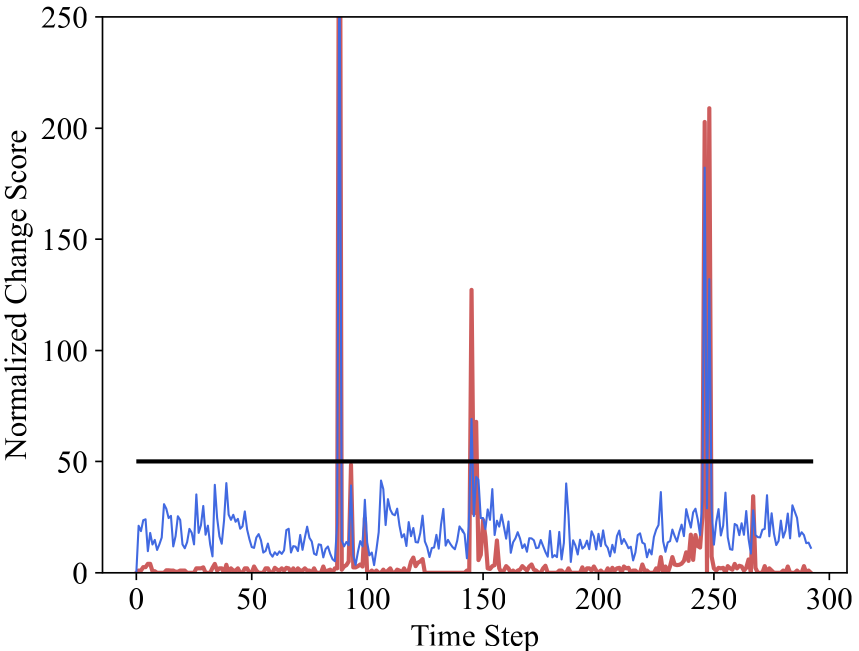}
    \caption{An example of the change scores $s_{\mathrm{pos}}$ (in red) and $s_{\mathrm{feat}}$ (in blue) in WrapCandy task. $s_{\mathrm{feat}}$ is multiplied by 1000 so that both scores share the same scale. The horizontal lines indicate the thresholds $\theta_{\mathrm{pos}} = 50$ and $\theta_{\mathrm{feat}} = 0.05$.}
    \label{change_scores}
\end{figure}

\subsubsection{Change Detection}
To detect abrupt changes caused by gaze transitions, we compute the difference between the current and previous time steps for the filtered gaze position $\hat{g}_t$ and visual feature $f_t$.
\begin{align}
    s^{\mathrm{pos}}_t = \bigl\lVert\hat{g}_t - \hat{g}_{t-1}\bigr\rVert,
\end{align}
\begin{align}
    s^{\mathrm{feat}}_t = -\frac{\log\bigl({f^\mathrm{left}_{t-1}}^{T}f^\mathrm{left}_{t} + 1\bigr) + \log\bigl({f^\mathrm{right}_{t-1}}^{T}f^\mathrm{right}_{t} + 1\bigr)}{2} \notag \\+ \log 2,
\end{align}
where $s^{\mathrm{pos}}_t$ and $s^{\mathrm{feat}}_t$ represent the change scores for the gaze position and the visual features, respectively. Specifically, $s^{\mathrm{pos}}_t$ is the Euclidean distance between the two 4-dimensional vectors representing the left and right gaze positions. Meanwhile, $s^{\mathrm{feat}}_t$ is derived by calculating  the inner products between the corresponding gaze feature vectors in consecutive time steps, normalized to ensure non-negativity, and then averaged for the left and right features.

Fig. \ref{change_scores} illustrates a typical time series of the change scores from a demonstration.
In this example, each sub-task corresponds to one of four temporal segments, marked by sudden spikes in the scores.
To detect these transitions, we introduce the thresholds $\theta_{\mathrm{pos}}$ and $\theta_{\mathrm{feat}}$ as hyperparameters for the two change scores. Any time step is identified as a change point $c_i \in C$ if both scores exceed their respective thresholds.

\subsection{Refinement Based on Decomposition Consistency}
In machine learning–based imitation learning, particularly methods using multiple demonstrations, all demonstrations of the same task typically comprise a consistent set of sub-tasks. 
This consistency allows for further refinement to improve the consistency of task decomposition across all these demonstrations.
To achieve this, we introduce a refinement procedure, $\displaystyle C \rightarrow C^*$, to ensure that all demonstrations of the same task share as consistent a number of sub-tasks as possible. 
The algorithm for this process is summarized in Algorithm \ref{alg:refinement}.

First, we determine the number of change points $\lvert C \rvert$ obtained via the gaze transition detection (Section \ref{gaze-transition-detection}) for every demonstration and denote the most frequent value by $s$. We assume $s$ to be the number of sub-tasks for the task and automatically adjust the thresholds $\theta_{\mathrm{pos}}$ and $\theta_{\mathrm{feat}}$ so that the number of segments in each demonstration matches $s$. 
These thresholds are adjusted individually for each demonstration.
If a demonstration has fewer than $s$ segments, both thresholds are reduced by 1\% per iteration. Conversely, if the count exceeds $s$, the thresholds are increased by 1\% per iteration. If a suitable threshold combination cannot be found within a predefined number of iterations, the demonstration is considered a detection failure and excluded from the training data.

Finally, the refined set of change points, $C^*$, obtained by this process is taken as the sub-task boundary for each demonstration, enabling us to divide as many demonstrations as possible into a consistent number of sub-tasks.

\begin{algorithm}[t]
\caption{Task Decomposition Refinement Process}
\label{alg:refinement}
\begin{algorithmic}[1]
    \Statex
    \textbf{Given:} A set of demonstrations for the same task:
    \[
      D = \{(s_{t}^{\mathrm{pos}},\,s_{t}^{\mathrm{feat}})^{(n)}\}_{n=1}^{N}, \quad t \in [0,\,T^{(n)}],
    \]
    where \(N\) is the number of demonstrations and \(T^{(n)}\) is the total time steps in the \(n\)-th demonstration.

    \Statex
    \textbf{Notation:} Let the function \(\textit{detect}(\cdot)\) perform change point detection:
    \[
      C \;\leftarrow\; \textit{detect}\bigl(\{s_{t}^{\mathrm{pos}}\},\,\{s_{t}^{\mathrm{feat}}\},\,\theta_{\mathrm{pos}},\,\theta_{\mathrm{feat}}\bigr).
    \]

    \Statex
    \textbf{Initialize:} $\theta_{\mathrm{pos}}, \theta_{\mathrm{feat}} \;\leftarrow\; \text{default values}$

    \For{\(n = 1\) to \(N\)}
        \State 
          \(C^{(n)} \;\leftarrow\; \textit{detect}\bigl(\{s_{t}^{\mathrm{pos}}\}^{(n)},\,\{s_{t}^{\mathrm{feat}}\}^{(n)},\,\theta_{\mathrm{pos}},\,\theta_{\mathrm{feat}}\bigr)\)
    \EndFor

    \State \( s \;\leftarrow\; \text{mode}(\{|C^{(n)}|\}_{n=1}^{N}) \ \ // 
 \ \text{fixed value}\)

    \For{\(n = 1\) to \(N\)}
        \State $\theta_{\mathrm{pos}}, \theta_{\mathrm{feat}} \;\leftarrow\; \text{default values}$
        \While{\(\bigl|C^{(n)}\bigr| < s\)}
            \State 
              \(\theta_{\mathrm{pos}},\,\theta_{\mathrm{feat}} 
                \;\leftarrow\; 0.99\,\theta_{\mathrm{pos}},\; 0.99\,\theta_{\mathrm{feat}}\)
            \State 
              \(C^{(n)}
                \leftarrow \textit{detect}(\{s_{t}^{\mathrm{pos}}\}^{(n)},\{s_{t}^{\mathrm{feat}}\}^{(n)},\theta_{\mathrm{pos}},\theta_{\mathrm{feat}})\)
        \EndWhile

        \While{\(\bigl|C^{(n)}\bigr| > s\)}
            \State 
              \(\theta_{\mathrm{pos}},\,\theta_{\mathrm{feat}}
                \;\leftarrow\; 1.01\,\theta_{\mathrm{pos}},\; 1.01\,\theta_{\mathrm{feat}}\)
            
            \State 
              \(C^{(n)}
                \leftarrow \textit{detect}(\{s_{t}^{\mathrm{pos}}\}^{(n)},\{s_{t}^{\mathrm{feat}}\}^{(n)},\theta_{\mathrm{pos}},\theta_{\mathrm{feat}})\)
        \EndWhile
        \State $C^{*(n)} \;\leftarrow\; C^{(n)}$
    \EndFor
    
    \Statex 
    \textbf{Return:} $\{C^{*(n)}\}_{n=1}^{N}$
\end{algorithmic}
\end{algorithm}

\section{EXPERIMENTS}
\subsection{Robot System}
A dual-arm robotic system that was previously developed for imitation learning through teleoperation is used in this study \cite{Kim2020}. A human teleoperator remotely controlled the robot by viewing its environment through an HMD (Vive Pro Eye, HTC Corporation) equipped with an integrated eye tracker, which simultaneously recorded the operator’s eye gaze. The system is configurable by a physical dual-arm robot, a pair of UR5 robots from Universal Robots Inc., or a simulated counterpart, enabling the collection of expert demonstrations in both real and virtual settings  \cite{Hamano2022}. The recorded gaze data were provided as pixel coordinates corresponding to the images displayed on the HMD.

This study evaluates the task decomposition method using an expert demonstration dataset collected through this system. The demonstrations were recorded at 10 Hz, capturing synchronized left and right image streams along with corresponding gaze data, all of which were used in our experiments.

\subsection{Task Setup}
We evaluated the proposed methods on three object manipulation tasks using the physical robot system.
For each task, we collected approximately 100 demonstrations as follows:
\begin{itemize}
    \item \textbf{WrapCandy (real robot, 64 demos)}: A green candy and a brown handkerchief are placed on a table covered with a green tablecloth. The position and orientation of the two objects are randomized on the table. The robot grasps the candy and places it in the center of the handkerchief. The robot then picks up the (top) left corner of the handkerchief and folds it to wrap the candy. This task features a relatively long horizon.
    \item \textbf{MoveFlask (real robot, 106 demos)}: A flask stand and a flask placed on the table. The initial position of the flask on the stand is randomized. The robot grasps the flask and moves it to a different location on the flask stand. This task is characterized by relatively small movements on the flask stand (13 cm x 7 cm), making it difficult to distinguish individual sub-tasks. Unlike the other two tasks, the demonstrations for this task were collected by an operator other than the author.
    \item \textbf{OpenCap (real robot, 110 demos)}: A labeled plastic bottle is placed on the table without a tablecloth. The position and orientation of the bottle are randomized. The robot holds the bottle in place with its right arm and removes the cap with its left arm. This task is characterized by the use of dual arms. 
\end{itemize}

\subsection{Task Decomposition Results}
\begin{figure}
    \begin{tabular}{c}
        \begin{minipage}[c]{0.95\linewidth}
            \centering
            \includegraphics[width=\linewidth]{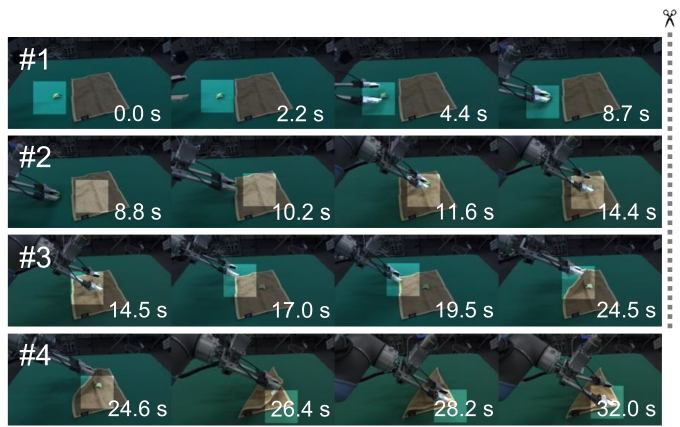}
            \subcaption{}
        \end{minipage}
        \\
        \begin{minipage}[c]{0.95\linewidth}
            \centering
            \includegraphics[width=\linewidth]{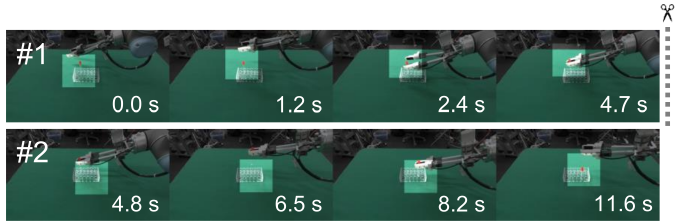}
            \subcaption{}
        \end{minipage}
        \\
        \begin{minipage}[c]{0.95\linewidth}
            \centering
            \includegraphics[width=\linewidth]{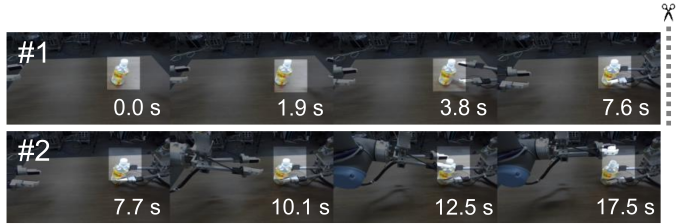}
            \subcaption{}
        \end{minipage}
    \end{tabular}
    \caption{Typical task decomposition results using the proposed method for the tasks of WrapCandy (a), MoveFlask (b), and OpenCap (c). The unshaded square areas represent the images cropped around the gaze positions.}
    \label{seg-results}
\end{figure}

For each task, we performed task decomposition across all demonstrations in the dataset, using the same hyperparameters across all demonstrations and tasks: $w=20, b=256 \;(\text{overall image size: }1280\times720), \theta_{\mathrm{pos}} = 50, \theta_{\mathrm{feat}}=0.03$.
Fig. \ref{seg-results} presents a representative example of segmentation results for each task.

In the WrapCandy task, the series of object manipulations was divided into four sub-tasks: (1) picking up the candy, (2) placing the candy at the center of the towel, (3) grasping the upper-left corner of the towel, and (4) folding the towel.  
In the MoveFlask task, the demonstration was divided into two sub-tasks: (1) picking up the flask and (2) moving the flask to a different position.  
In the OpenCap task, the demonstration was divided into two sub-tasks: (1) holding the plastic bottle with the right end-effector and (2) removing the cap with the left end-effector. These segmentations align with human intuition, effectively capturing meaningful task transitions.

A closer examination of the states before and after each segmentation revealed two common features across the tasks:
\begin{itemize}
    \item Each sub-task began when the robot (released the previous object and) started reaching for the next relevant object.  
    \item Each sub-task ended once the action’s completion was visually confirmed. (For instance, in a grasping action, the sub-task did not end when the hand merely contacted the object but rather when the object was lifted slightly off the table, confirming a successful grasp.)
\end{itemize}

\subsection{Consistency of Task Decomposition} 
In this section, we assessed the consistency of the resulting task decomposition.
Table \ref{tab:ablation} shows the number of demonstrations classified as “minority data,” where the segmentation timings or the number of sub-tasks differed from the most common (majority) decomposition for each task.
To evaluate the impact of different components of our approach, we compared the proposed method against several ablation settings: (1) without the refinement process (\textbf{w/o refine}), (2) detecting changes based solely on the gaze position $\hat{g}_t$ (\textbf{w/o feat}), (3) detecting changes based solely on visual features around the gaze $f_t$ (\textbf{w/o pos}), and (4) omitting the refinement process from these two settings (\textbf{w/o feat, refine} and \textbf{w/o pos, refine}).

\begin{table}[tb]
    \centering
    \caption{Number of minority demonstrations exhibiting task decomposition
    with different timing or segment counts from the majority in each task.}
    \label{tab:ablation}
    \resizebox{\linewidth}{!}{
    \begin{tabular}{lcccc}
        \toprule
        Method & \textbf{WrapCandy} & \textbf{MoveFlask} & \textbf{OpenCap} & \textbf{Total} \\
        \midrule
        Ours                & \textbf{0} / 64    & \textbf{1} / 106  & \textbf{0} / 110  & \textbf{1} / 280 \\
        - w/o refine        & 1         & 7        & 39       &  47              \\
        - w/o feat          & \textbf{0}         & \textbf{1}        & \textbf{0}        & \textbf{1}       \\
        - w/o feat, refine  & 1         & 6        & 22       & 29               \\
        - w/o pos           &\textbf{0}         & \textbf{1}        & 24       & 25               \\
        - w/o pos, refine   & 34        & 5        & 55       & 94               \\
        \bottomrule
    \end{tabular}
    }
\end{table}

\begin{figure*}[t]
    \centering
    \begin{tabular}{ccc}
        \begin{minipage}[]{0.31\linewidth}
            \centering
            \includegraphics[width=\linewidth]{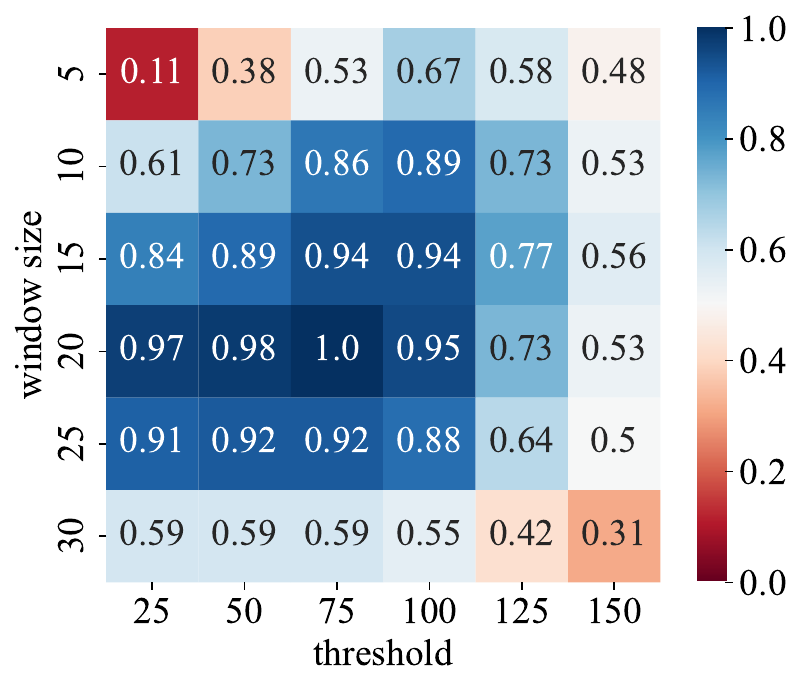}
            \label{hypara-heatmap-a}
        \end{minipage}
        \begin{minipage}[]{0.31\linewidth}
            \centering
            \includegraphics[width=\linewidth]{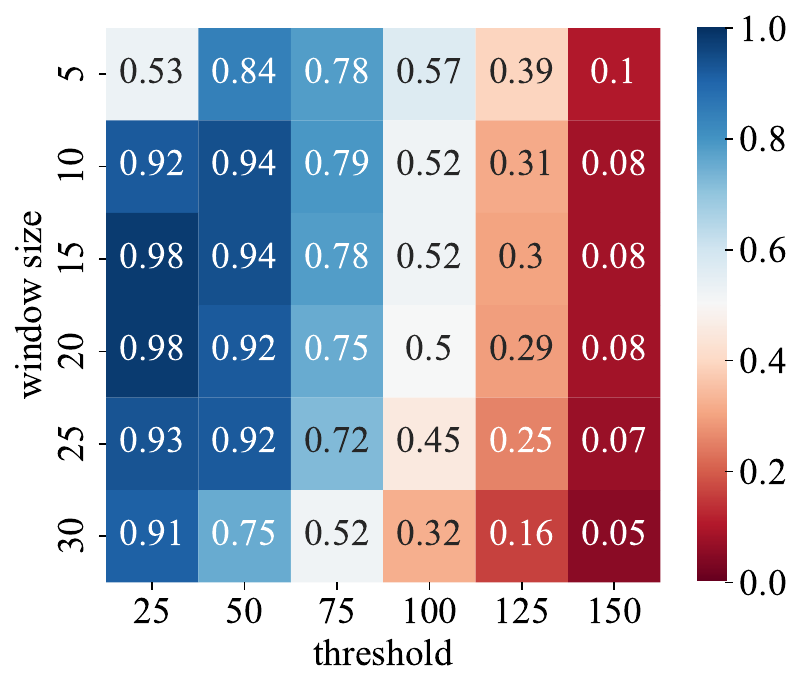}
        \label{hypara-heatmap-b}
        \end{minipage}
        \begin{minipage}[]{0.31\linewidth}
            \centering
            \includegraphics[width=\linewidth]{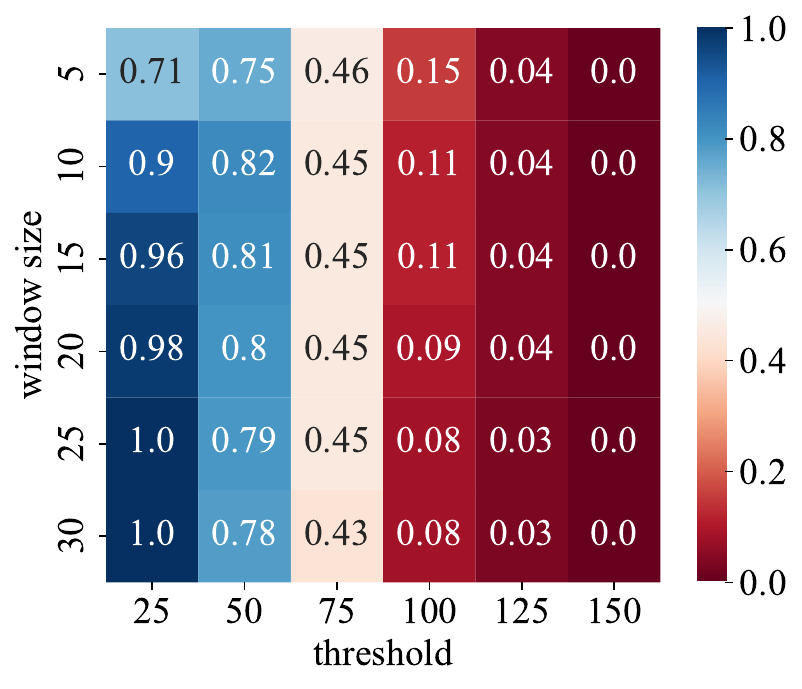}
        \label{hypara-heatmap-b}
        \end{minipage}
        \vspace{0.01\linewidth} \\
        \begin{minipage}[]{0.31\linewidth}
            \centering
            \includegraphics[width=\linewidth]{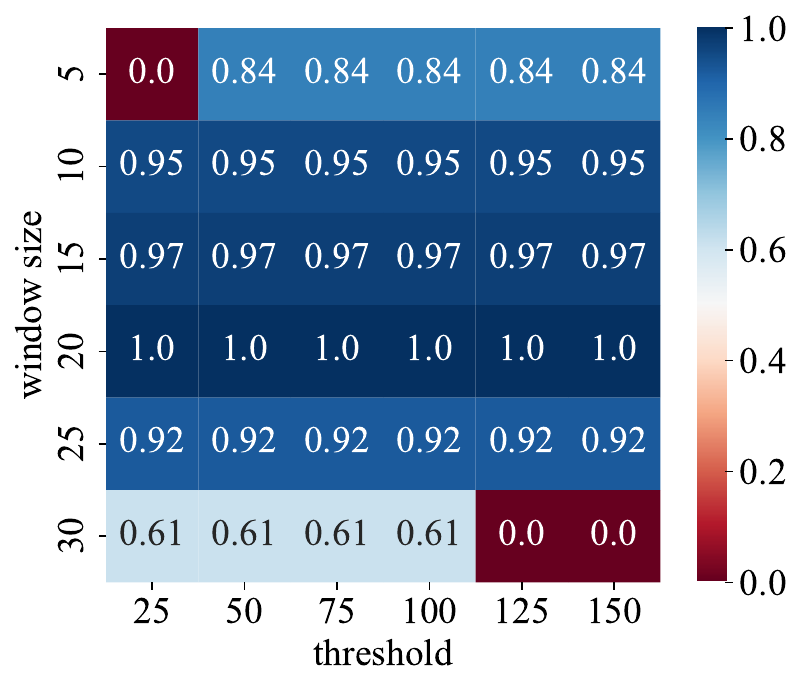}
            \subcaption{WrapCandy}
            \label{hypara-heatmap-a}
        \end{minipage}
        \begin{minipage}[]{0.31\linewidth}
            \centering
            \includegraphics[width=\linewidth]{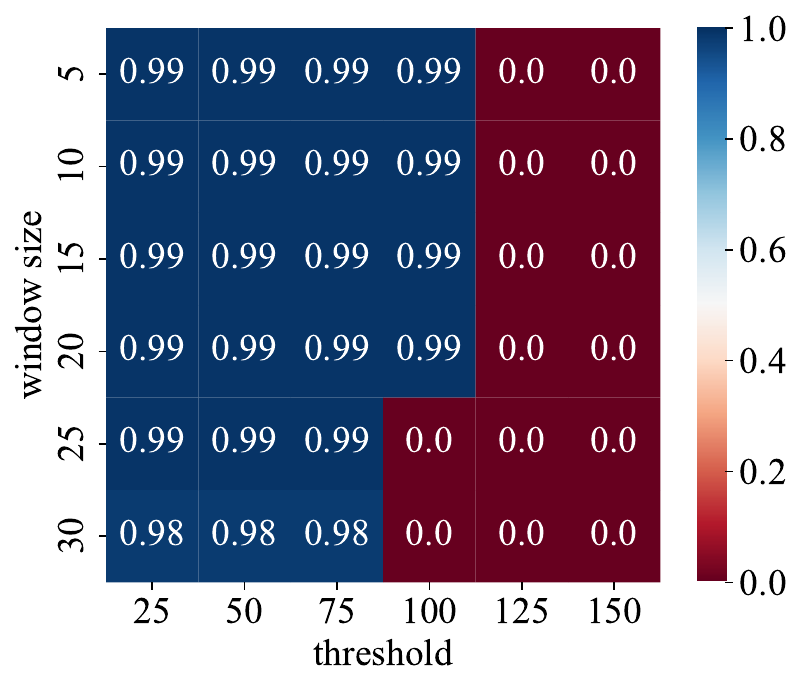}
            \subcaption{MoveFlask}
        \label{hypara-heatmap-b}
        \end{minipage}
        \begin{minipage}[]{0.31\linewidth}
            \centering
            \includegraphics[width=\linewidth]{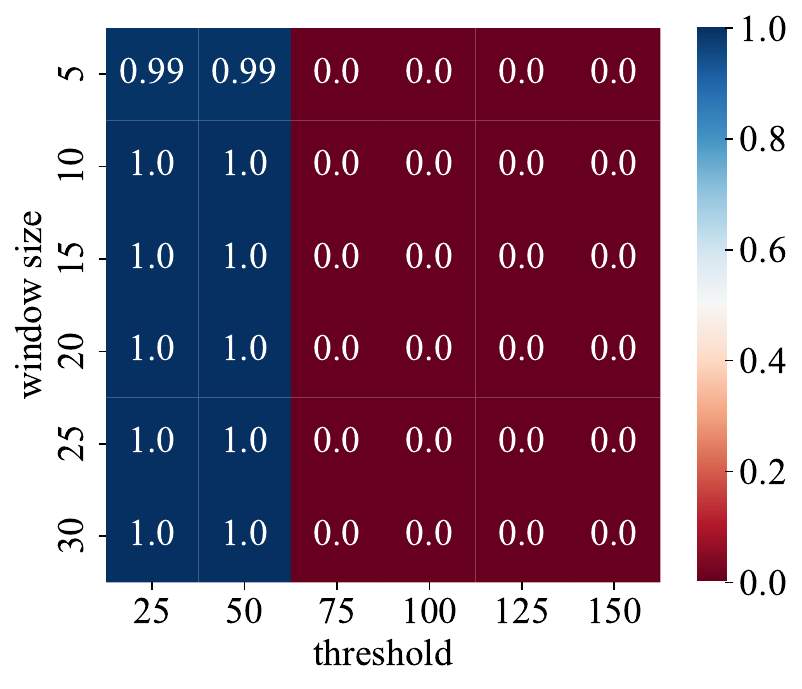}
            \subcaption{OpenCap}
        \label{hypara-heatmap-b}
        \end{minipage}
    \end{tabular}
    \caption{Success rate of decomposed demonstrations with varying hyperparameters for each task. (top) Without the refinement process. (bottom) With the refinement process.}
    \label{hypara-heatmap}
\end{figure*}

As illustrated in Table \ref{tab:ablation}, applying the refinement process significantly improved the consistency of task segmentation across all tasks.
The largest difference in segmentation performance was observed in the OpenCap task. This discrepancy arose because the change in gaze information between holding the bottle with the right end-effector and grasping the cap with the left was less pronounced compared to the other two tasks, making segmentation more challenging without refinement.
Due to this smaller change, some demonstrations were missed under the fixed threshold described earlier.

When the gaze position was excluded (i.e., w/o pos, w/o pos, refine), the number of minority cases increased substantially compared to settings that included the gaze position, even after the refinement process. 
This occurred because the visual features alone had a lower signal-to-noise ratio than gaze position. 
Consequently, over-segmentation was observed when the gaze shifted slightly during an action, whereas segmentation was sometimes missed when the gaze displacement was too small (e.g., when objects were close together). 
Additionally, in some demonstrations, change detection was triggered whenever the end-effector or a grasped object entered the cropped image region, leading to over-segmentation.

In the MoveFlask task, only one demonstration remained segmented incorrectly even after the refinement. 
In this case, the teleoperator shifted their gaze to the destination before grasping the flask.

Overall, these results indicate that incorporating the gaze position and applying the refinement process consistently achieved the decomposition of nearly all demonstrations. Notably, in the tasks we evaluated, adding visual features did not enhance segmentation performance.

\subsection{Hyperparameter Robustness}
Ensuring consistent task decomposition results remain consistent despite changes in hyperparameters is critical for the generality and robustness of the proposed method. In this section, we analyze how variations in the hyperparameters $w, b, \theta_{\mathrm{pos}}, \theta_{\mathrm{feat}}$ impact the number of correctly segmented demonstrations.

From our previous experiment (Table \ref{tab:ablation}), we confirmed that when gaze position was employed for change detection, the presence or absence of gaze feature data had no significant impact on the segmentation results. Therefore, in this experiment, we performed task decomposition using only the gaze position. Under this condition, the relevant hyperparameters are $w$ and $\theta_{\mathrm{pos}}$. Specifically, $w$ determines the filtering strength of the median filter applied to the gaze position, while $\theta_{\mathrm{pos}}$ determines the sensitivity for detecting gaze transitions.

Fig. \ref{hypara-heatmap} shows the number of demonstrations that were correctly segmented when $w$ was set to $\{5, 10, 15, 20, 25, 30\}$ and $\theta_{\mathrm{pos}}$ was set to $\{25, 50, 75, 100, 125, 150\}$.
Without the refinement process (Fig. \ref{hypara-heatmap} top), incorrect decompositions fluctuated gradually as the hyperparameters changed. However, with refinement processing applied, decomposition was successfully achieved for almost all demonstrations across the broad parameter range (Fig. \ref{hypara-heatmap} bottom). These results demonstrate the robustness of the proposed method, making it adaptable to different tasks and robotic systems.

In cases where incorrect task decomposition persisted after the refinement process, including instances where the number of successfully segmented demonstrations was lower with refinement (Fig. \ref{hypara-heatmap} top) than without it (Fig. \ref{hypara-heatmap} bottom), decomposition errors resulting from an incorrect number of segments had already become predominant beforehand. 
Consequently, instead of fulfilling its intended purpose, the refinement process altered demonstrations that were initially correctly segmented, leading to an incorrect number of sub-tasks.
Furthermore, setting the window width $w$ to 25 or higher caused the influence of median filtering to become excessively strong. As a result, in the WrapCandy task, it became impossible to detect the required number of segmentation points during the refinement process, regardless of how much the threshold $\theta_{\mathrm{pos}}$ was lowered. 
Compared to the WrapCandy task, the MoveFlask and OpenCap tasks involve more delicate end-effector and gaze movements. In these tasks, gaze transitions could not be detected in the majority of demonstrations when the threshold was increased beyond certain levels (125 or higher in MoveFlask and 75 or higher in OpenCap).

\subsection{Application to Imitation Learning}
We integrated our proposed task decomposition method into an imitation learning framework. 
The framework proposed in \cite{Takizawa2025b} requires a sequence of actions to be decomposed into sub-tasks as a prerequisite for subsequent temporal and spatial segmentation. This segmentation enhances the reusability of learned skills in previously unseen object positions and end-effector poses.
To automate the sub-task decomposition process, we incorporated our gaze-based task decomposition method.
The policy model trained using demonstrations automatically segmented by our method achieved a higher task success rate in previously unseen object positions and end-effector poses compared with conventional imitation learning models.
For more details, please refer to \cite{Takizawa2025b}.

\section{CONCLUSIONS}
This study proposed a practical method for achieving consistent and robust task decomposition for robot manipulation in a recent imitation learning framework.
Our key finding is that the desired task decomposition was successfully achieved for all demonstrations by adjusting the threshold in the refinement process.
This result highlights the fundamental effectiveness of utilizing gaze transitions for task decomposition.

However, the proposed method has several limitations.
Based on the characteristics of the current imitation learning dataset, we assumed that all demonstrations for a given task consist of the same set of sub-tasks. 
However, in more practical settings, the actions taken to accomplish the same task may consist of a different number of sub-tasks.
For example, when pouring water into a cup, one may either grasp the cup before pouring or pour without grasping it.
In even more general settings, demonstrations are no longer labeled by task, and a single demonstration may include multiple tasks, as seen in \textit{play data} \cite{Lynch2019}.
In such cases, while our gaze transition detection remains applicable, the refinement process cannot be directly applied.
To enhance the applicability of our approach, future work should focus on integrating the refinement process with sub-task categorization in an iterative manner. 
This would enable a more flexible and practical demonstration collection, ultimately contributing to a more robust and adaptable imitation learning framework for real-world deployment.





\section*{ACKNOWLEDGMENT}
We sincerely thank Izumi Karino for providing the demonstration data used in the experiments of this study.


\balance


\end{document}